*Article*

# RS-YOLOX: A High Precision Detector for Object Detection in Satellite Remote Sensing Images

**Lei Yang [1], Guowu Yuan [1,2,\*], Hao Zhou [1,2], Hongyu Liu [1], Jian Chen [1] and Hao Wu [1,2]**

[1] School of Information Science and Engineering, Yunnan University, Kunming 650504, China;
[2] Yunnan Key Laboratory of Intelligent Systems and Computing, Kunming 650504, China
yanglei@mail.ynu.edu.cn (L.Y.); gwyuan@ynu.edu.cn (G.Y.)
**\*** Correspondence: gwyuan@ynu.edu.cn (G. Y.)

**Abstract:** Automatic object detection by satellite remote sensing images is of great significance for resource exploration and natural disaster assessment. To solve existing problems in remote sensing image detection, this article proposes an improved YOLOX model for satellite remote sensing image automatic detection. This model is named RS-YOLOX. To strengthen the feature learning ability of the network, we used Efficient Channel Attention (ECA) in the backbone network of YOLOX and combined the Adaptively Spatial Feature Fusion (ASFF) with the neck network of YOLOX. To balance the numbers of positive and negative samples in training, we used the Varifocal Loss function. Finally, to obtain a high-performance remote sensing object detector, we combined the trained model with an open-source framework called Slicing Aided Hyper Inference (SAHI). This work evaluated models on three aerial remote sensing datasets (DOTA-v1.5, TGRS-HRRSD, and RSOD). Our comparative experiments demonstrate that our model has the highest accuracy in detecting objects in remote sensing image datasets.

## 1. Introduction

With the continuous expansion of remote sensing applications, remote sensing has been widely used in agriculture [1, 2], forestry [3], water conservancy [4], urban and rural construction [5], and earth science research [6, 7]. A crucial part of remote sensing applications is object detection in satellite remote sensing images, and it is of great significance for resource exploration, natural disaster assessment, etc. For example, Meng et al. [8] used a target detection method based on bags of visual word features to detect the leakage of seabed gas, their research could be used to explore natural gas hydrate and other resources. Liu et al. [9] proposed an improved Mask RCNN model to automatically identify landslides on the Eastern Tibet Plateau, so as to reduce the losses caused by landslides to a certain extent. In the field of object detection, the research on automatic detection algorithms for remote sensing images has always been important and challenging. However, due to the large size of remote sensing images, it is difficult to identify small objects, and the samples are unbalanced, which seriously affects the accuracy of recognition. Traditional detection algorithms are not effective in detecting objects in remote sensing images. This is because traditional detection and the recognition of remote sensing image targets are mainly based on manually extracting features, and the rich, diverse, and detailed information in remote sensing images means that a single feature described manually is inadequate at fully expressing the target characteristics and relies more on expert experience. In addition, machine learning based on probability and statistics usually requires complex feature description, and the feature representation learned on the basis of its shallow network structure is obviously insufficient in terms of performance and generalization ability when dealing with complex target detection problems.

In recent years, with the application and development of deep learning in the field of computer vision, many excellent object detection models based on deep learning methods have been proposed, for example, the two-stage model represented by the R-CNN series [10-12] and the one-stage model represented by SSD [13, 14] and YOLO [15-20]. These models have achieved impressive results in the detection of natural images. Researchers have introduced advanced detection algorithms into the application of remote sensing images, the goal being to overcome the difficulties in satellite remote sensing image detection. Yan et al. [21] used an improved Faster R-CNN model to detect tailing ponds in remote sensing images. Luz et al. [22] used an unsupervised model to detect fire-prone areas. Etten [23] proposed the YOLT model, which is an improved YOLOv2 model for satellite remote sensing image detection. Although these methods have achieved good results, they only improve one direction or detection of one category in remote sensing images and do not systematically solve the problems of remote sensing image detection.

Focusing on the problems of detecting objects in satellite remote sensing images, this article proposes a high-precision automatic detection method for satellite remote sensing images: RS-YOLOX. We obtained this model by improving the YOLOX model. We evaluated our proposed methods by using three aerial remote sensing datasets (DOTA-v1.5, TGRS-HRRSD, and RSOD). Our model achieved good results on these three datasets, outperforming all YOLO models in accuracy.

The novelty of this article is exemplified by the following:

- We propose a state-of-the-art YOLO model that surpasses all previous YOLO models in detection accuracy in satellite remote sensing images.
- We use Efficient Channel Attention (ECA) and the Adaptively Spatial Feature Fusion (ASFF) in YOLOX to improve the model's accuracy in detecting small objects in remote sensing images.
- We use a new loss function, Varifocal Loss function (VFL), which alleviates the problem of unbalanced training of positive and negative samples in model training.
- We combine the detection model with the Slicing Aided Hyper Inference framework (SAHI) to realize the direct detection of large-size remote sensing images.

The rest of this study is organized as follows: Section 2 introduces the current popular object detection methods and their application in remote sensing image detection and then introduces the YOLOX model that inspired this article. Section 3 focuses on our methods to improve YOLOX to address the problems of satellite remote sensing image detection. Section 4 presents the experimental results of our methods and our analysis of the results. Sections 5 and 6 discuss and conclude our work, respectivel.

**2. Related Work**

*2.1. Mainstream object detection algorithms*

The current mainstream object detection algorithms can be divided into two categories: two-stage and one-stage.

As the name suggests, the detection of the two-stage object detection algorithm is divided into two stages: (1) to extract the pre-selected boxes from the input image, (2) to use the convolutional neural network (CNN) to classify and regress. The most famous two-stage object detection algorithm is the R-CNN series [10-12]. Although the two-stage target detection algorithm can achieve high detection accuracy, the detection speed is slow and it is not suitable for real-time detection.

The one-stage algorithm directly extracts features from the input image and then calculates the category probability and position coordinates of the targets. Typical one-stage algorithms include SSD [13, 14] and YOLO series [15-20]. In the past, the detection accuracy of the one-stage algorithm was often lower than that of the two-stage algorithm. However, with the proposal of YOLOv3 [17], the one-stage model surpasses the two-stage model in accuracy, while maintaining high-speed detection.

*2.2. Object Detection in Remote Sensing Image*

In recent years, many researchers have used the latest two-stage algorithm or one-stage algorithm in satellite remote sensing images and made corresponding improvements to the algorithm.

Yan et al. [21] proposed an improved Faster R-CNN model to detect tailing ponds in remote sensing images. They used an attention mechanism in the Faster R-CNN, and the average precision (AP) of detecting tailing ponds reached 85.7%. However, the single dataset they used cannot reflect the generalization of the model.

Etten [23] proposed the YOLT model, an improved YOLOv2 model for satellite remote sensing image detection. This model uses a ResNet-like residual structure to connect the features of multiple layers to obtain a more fine-grained feature representation, improving the detection of small objects, but it does not solve the problem of the imbalanced number of positive and negative samples in remote sensing images.

Xie et al. [24] proposed applying the YOLOv5 algorithm to detect remote sensing images but did not optimize the YOLOv5 algorithm for the purpose.

*2.3. Attention Mechanisms in Vision Tasks*

The principle of the attention mechanism is to imitate the attentional thinking of human beings. It can shift the attention of the model learning to important informative regions and ignore irrelevant information. Attention algorithms used in computer vision include spatial domain attention, channel domain attention and mixed domain attention.

Spatial domain attention obtains attention weights by computing the correlation of each position in the feature map with all other positions. For example, STN (Spatial Transformer Networks) [25], which learns the spatial transformation parameters by performing reverse spatial transformation on the data, so as to reduce the influence of the deformation of the target on the picture, thereby making the recognition of the classification network more efficient.

Channel domain attention is to apply attention weights on channels to capture the dependencies between channels, a typical representative is SENet (Squeeze-and-Excitation Networks) [26], it uses global average pooling to process feature maps to generate per-channel statistics, then uses two fully connected layers to learn weighted values for each channel, and finally get a weighted feature map as the next layer network input.

Both spatial domain attention and channel domain attention algorithms are theoretically flawed. The former treats the features of each channel the same, ignoring the correlation between channels, while the latter performs global average pooling on each channel, local information is ignored. Therefore, some researchers proposed mixed domain attention algorithm, which combines the attention of the spatial domain and the channel domain. For example, CBAM [27], which uses a serial method, the feature map first passes through the channel attention module, and then passes through the spatial attention module.

All of the above three attention algorithms can be integrated with convolutional neural networks to complete visual tasks such as classification, detection, and segmentation. For example, Li et al. [28] fused channel attention with YOLOv5 to achieve disease detection in maize leaves, Wang et al. [29] introduced mixed attention into two-stream convolution network to realize human action recognition. Li et al. [30] proposed a new mixed attention algorithm and used it in the U-Net network to achieve cell segmentation.

It can be known from these studies that the proposal of the attention mechanism promotes the development of computer vision models to higher accuracy.

*2.4. YOLOX*

YOLOX is a high-performance object detection model proposed by Ge et al. [20], which is an upgrade of the YOLO series.

The YOLO series of models, starting from YOLOv3, have used the common three-part structure of an object detector: backbone for extracting features, neck for integrating features, and head for implementing detection. YOLOX continues to use this network structure (Figure 1).

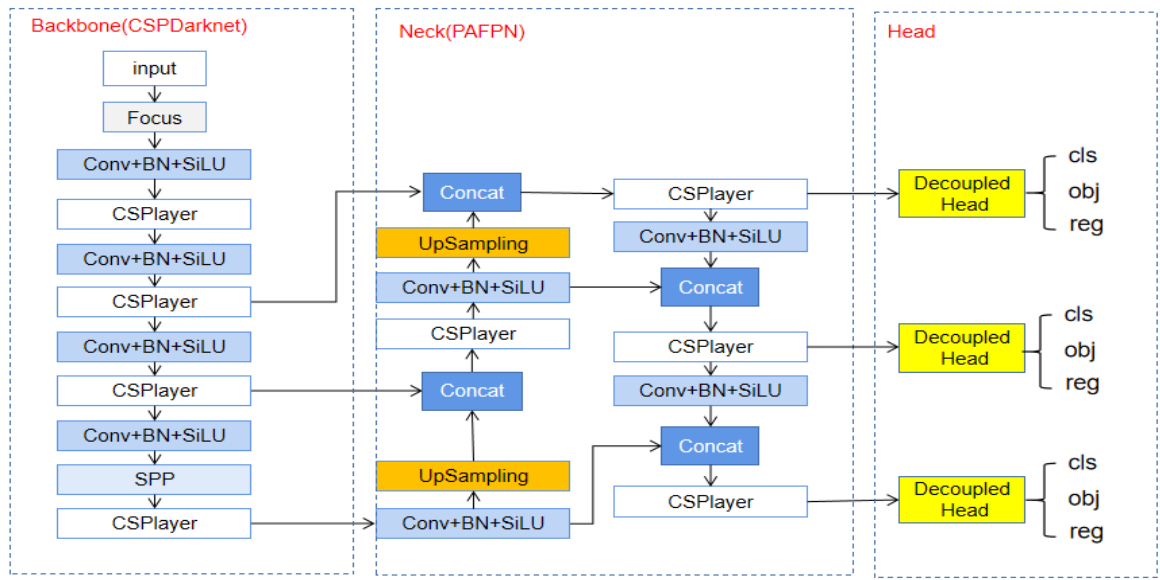

**Figure 1**. Structure of YOLOX.

As shown in Figure 1, to increase the number of channels, YOLOX uses the Focus module from YOLOv5. In addition, it combines with the feature extraction network CSPDarkNet from YOLOv4 as the backbone network, followed by the SPP module [31] for multi-scale training. The neck part of YOLOX uses the PAFPN [32] structure to realize the integration of multi-scale features. After the input image is processed by the backbone network and the neck network, it is divided into three scales, 13×13, 26×26, and 52×52, to predict large objects, medium objects, and small objects, respectively. Finally, the features of the three scales are respectively entered into the decoupled detection head to complete the detection.

The biggest improvement is that YOLOX uses anchor-free mode [33-35] and a decoupled head [36]. Anchor-free mode gets rid of the calculation required when using the preset anchors and greatly reduces the number of parameters. The coupled detection head couples the classification and regression of the prediction boxes, which will affect the detection performance. Therefore, YOLOX decouples classification and regression (i.e., separates them into two branches) and adds a branch for IOU calculation.

Although YOLOX is already an excellent detection model, there are some problems in using it to detect remote sensing images. We summarize these questions into the following three points:

- Satellite remote sensing images contain rich, detailed information and many small targets, making it inconducive for the detection model to extract useful target information. If we only rely on ordinary convolutional structures to extract features, some key information may not be valued by model learning. Therefore, we need to introduce a new attention module and feature fusion module to enhance the feature processing capability of the convolutional network in YOLOX.
- In object detection, the target information is the positive sample and the background information is the negative sample. Due to the complex background of satellite remote sensing images, the numbers of positive and negative samples are unbalanced, which seriously affects the accuracy of model detection. The loss function used by YOLOX is the Binary Cross Entropy (BCE) function, which is a traditional loss function that cannot balance the numbers of positive and negative samples.
- Many satellite remote sensing images have a size of tens of millions of pixels, and target detection models, including YOLOX, have poor detection performance where large-size images are concerned.

**3. Proposed Methods**

To overcome the problems described in Section 2.3, we used the following methods to improve YOLOX:

- We added the ECA modules [37] to the backbone network of YOLOX and connected the ASFF structure [38] after the neck network. This can enhance the feature extraction and fusion of the network, thereby improving the accuracy of the model in detecting small objects.
- We replaced the BCE function with the VFL function [39], which not only balances the numbers of positive and negative samples but also highlights the key samples in the positive samples.
- In the detection of large-size images, we used the SAHI framework [40], which can cooperate with the target detection model to achieve high-precision detection of large-size images.

*3.1. Network structure*

Figure 2 displays the network structure of our improved YOLOX model. we added the ECA modules to the back of the backbone network to strengthen the learning of the captured features and the ASFF to the back of the neck network to enhance the feature fusion. The VFL function is used in the decoupling heads as the loss function in training.

Finally, to improve the detection performance of the model for high-resolution remote sensing images, we combined the improved YOLOX model with the SAHI framework. We named the improved model RS-YOLOX. Figure 3 displays the detection process of this model. After the SAHI framework slices the high-resolution image, it is input into our target detection model. After detection is complete, all the slices are merged into a complete detection result map.

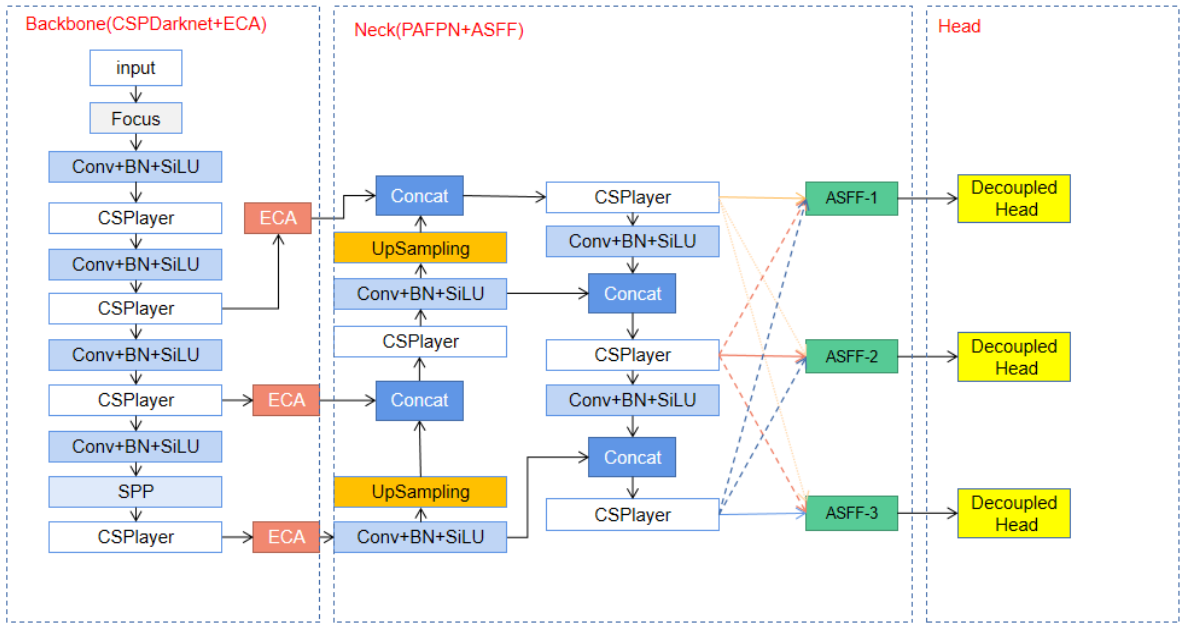

**Figure 2.** Structure of our improved YOLOX model.

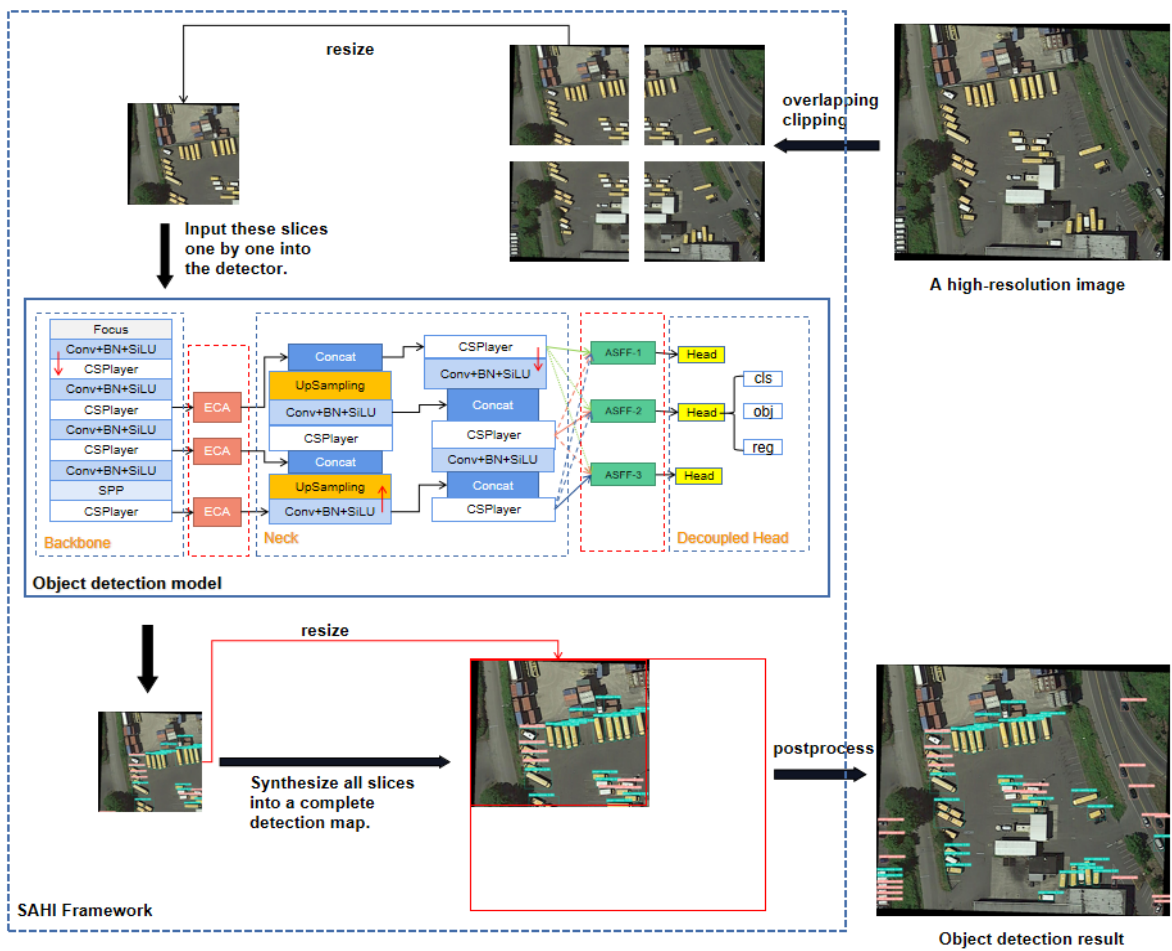

**Figure 3.** Detection process of RS-YOLOX.

*3.2. Efficient Channel Attention (ECA)*

Attention mechanisms in computer vision can ensure that models pay more attention to important information and selectively forget unimportant information or dis-turbing information that affects judgment. Therefore, object detection models can capture better features by using attention mechanisms, improving the performance of the model in detecting small objects. The ECA blocks [37] adopted in this article are an improvement over SE blocks [26].

SE is the beginning of channel attention mechanisms in computer vision. It strengthens important features by modeling the feature correlation between channels, thereby improving the detection performance of the model. Figure 4 shows the structure of the SE block.

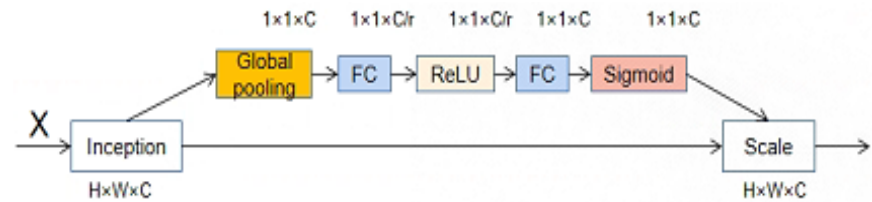

**Figure 4.** SE block structure.

The SE block uses two fully connected layers to capture nonlinear cross-channel interaction information. The parameters of the two fully connected layers are learned by network training. These parameters act on the feature map and can extract more important features. To reduce the complexity of the model, the first fully connected layer reduces the number of channels, which makes the model more lightweight but weakens the connection between the channel and the weight vector and is not good for capturing the interrelationships between channels. To address this shortcoming, the ECA block cancels the operation of dimensionality reduction and uses one-dimensional (1D) convolution, which not only reduces the number of parameters but also improves the detection accuracy of the model. ECA removes the fully connected layer in the SE block and directly learns through a weight-sharing 1D convolution on the features after global average pooling (GAP). One-dimensional convolution involves the hyperparameter k, which is the size of the convolution kernel and represents the coverage of local cross-channel interactions. ECA attempts to capture inter-channel interactions using each channel of feature maps and its k neighbors. Figure 5 displays the structure of an ECA block.

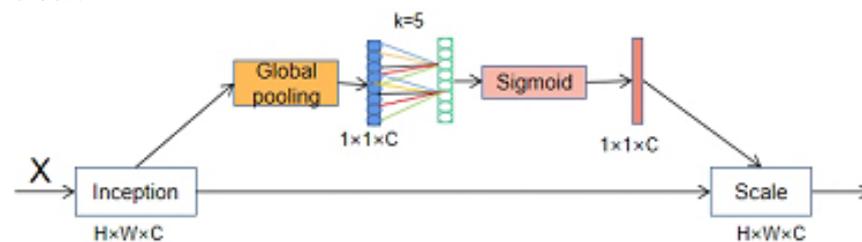

**Figure 5.** ECA block structure.

In Figure 5, the weights of 1D convolution are interleaved, realizing the function of mutual cross-channel. Weights exist in groups, and the number of weights in a group depends on the size of the convolution kernel. The values of weights of each group are shared, which greatly reduces the number of parameters. The attention of SE is calculated through two fully connected layers, while the attention of ECA is calculated through k nearest-neighbor channels. The size of k obviously affects the efficiency of ECA block calculation. In fact, k is related to the number of channels C; the nonlinear mapping relationship between them can be described by an exponential function. This can be formulated as:

$$C = \phi(k) \approx \exp(\gamma * k - b) \tag{1}$$

The number of channels is usually an exponential multiple of 2, so k can be calculated by the following formula:

$$k = \Psi(C) = |\frac{\log_2 C}{\gamma} + \frac{b}{\gamma}|_{odd} \qquad (2)$$

The structural design of ECA avoids dimensionality reduction, realizes cross-channel information interaction, and reduces the amount of calculation. It is an effective ultra-lightweight attention mechanism module. We added ECA blocks to the backbone network of YOLOX so that the features extracted by the backbone network were enhanced on the attention weight matrix.

*3.3. Adaptively Spatial Feature Fusion (ASFF)*

In addition to using ECA to enhance the features captured by the backbone network, this article also optimizes the feature fusion of the neck network, adding an effective feature fusion algorithm: ASFF [38].

The neck network of YOLOX adopts the structure of a multi-level feature pyramid: PAFPN. The structure enables the output of the neck network to combine both the location information of the shallow network and the semantic information of the deep network, thereby achieving better regression and classification results. ASFF can be connected to PAFPN to perform feature fusion on the feature map processed by PAFPN, which can fuse similar features together, improving the detection performance. Figure 6 displays the structures of PAFPN and ASFF.

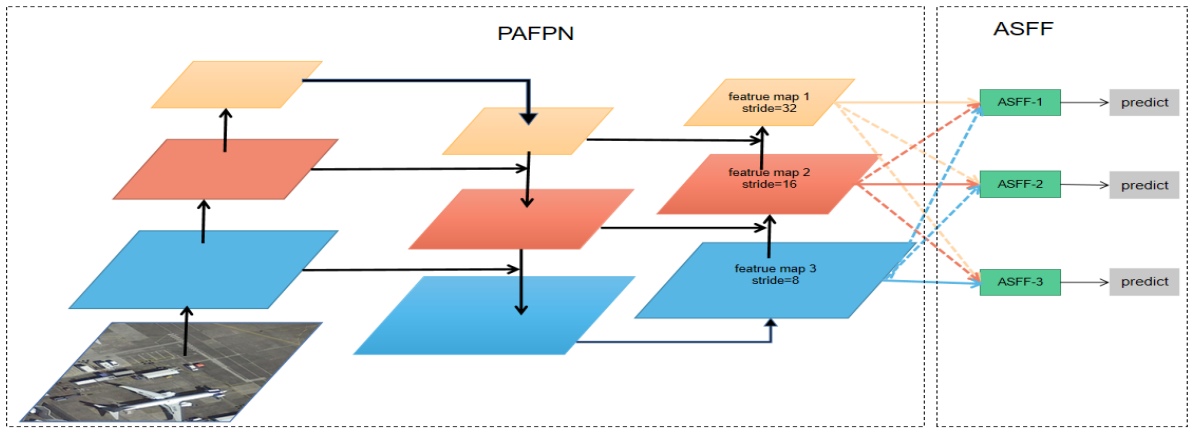

**Figure. 6** Structure of PAFPN and ASFF.

As shown in Figure 6, the image will output three feature maps through PAFPN. We number them from top to bottom: feature map $X^1$, feature map $X^2$, and feature map $X^3$. Arrows in different colors represent feature transfer in different feature maps.

$X^1$ has the largest receptive field and is suitable for detecting large objects in the picture; $X^3$ has the smallest receptive field and is suitable for detecting small objects; and $X^2$ has a receptive field in between, so it is used to detect medium-sized objects. ASFF takes the output ($X^1$, $X^2$, $X^3$) of the PAFPN network as input for the fusion of three feature maps. The calculation process of ASFF includes two parts: feature resizing and adaptive fusion.

Feature resizing involves scaling a feature map to the size of the feature map to be fused in order to ensure that the size of the feature map during feature fusion remains the same. We use $X^{1\rightarrow 3}$ to represent the image after $X^1$ is scaled to the size of $X^3$. Adaptive fusion is used to train three weight maps, α, β, and γ, through the network and multiply them with the input feature maps.

Taking the input of ASFF-3 in Figure 6 as an example: (1) Perform feature scaling on $X^1$, $X^2$, and $X^3$ to obtain $X^{1->3}$, $X^{2->3}$, and $X^{3->3}$, respectively. (2) Use the network to train three weight maps ($\alpha$, $\beta$, and $\gamma$) with the same size as $X^{1->3}$, $X^{2->3}$, and $X^{3->3}$ and multiply each weight map and each input point by point to determine the activation and suppression of each pixel in the input feature map X to achieve feature fusion. (3) Add the three matrices obtained after point-by-point multiplication to obtain the output of the feature fusion result of small objects in ASFF. The formula for the above calculation is as follows:

$$\text{ASFF-3} = X^{1->3} \otimes \alpha^3 + X^{2->3} \otimes \beta^3 + X^{3->3} \otimes \gamma^3 \tag{3}$$

Through such calculation, the features of small objects are enriched in the feature map $X^3$ used to detect small objects (the features of the small objects in the feature map $X^1$ and the feature map $X^2$ are fused). At the same time, the activation values of the features of large objects and medium objects in the feature map $X^3$ are filtered out so that the model is more focused on detecting small objects. The branches of ASFF-1 and ASFF-2 are also based on the same principle. They both integrate the features of their own concern in the three feature maps and suppress unimportant features, which improves the accuracy of object detection. The datasets of satellite remote sensing images contain a large number of small objects. Using ASFF, the features of these small objects can be fused as much as possible, improving the detection of small objects.

*3.4. Varifocal Loss function (VFL)*

The YOLOX model divides the loss function into classification loss, regression loss, and confidence loss, and the total loss function is the sum of the three. YOLOX uses the BCE function to calculate the classification loss and the confidence loss of the objects, using IoU or GIoU as the loss for prediction boxes regression. To solve the problem of unbalanced samples in satellite remote sensing images, this article replaces the BCE function with the VFL function [39]. VFL is an optimization of Focal Loss [41], and Focal Loss is an optimization of BCE. The loss function of BCE is as follows:

$$\text{BCE} = -\log(P_t) = \begin{cases} -\log(\hat{y}), & y = 1 \\ -\log(1 - \hat{y}), & y = 0 \end{cases} \tag{4}$$

In the above formula, y = 1 represents a positive sample and y = 0 represents a negative sample. As a classic loss function, BCE is widely used in classification and detection tasks but it does not suppress the number of negative samples and so cannot solve the problem of serious imbalance between positive and negative samples.

Focal Loss improves BCE, and its calculation formula is as follows:

$$\text{FL} = -\alpha_t(1 - P_t)^\gamma \log(P_t) = \begin{cases} -\alpha(1-\hat{y})^\gamma \log\hat{y}, & y = 1 \\ -(1-\alpha)\hat{y}^\gamma \log(1-\hat{y}), & y = 0 \end{cases} \tag{5}$$

$\alpha$ is used to adjust the weight of positive and negative samples, and *(1-Pt)* is used to reduce the influence of bad samples. The so-called bad samples refer to samples that are too easy to judge. If the sample is too easy to detect, its $P_t$ value will be high. VFL uses 1 to subtract this Pt value, thereby reducing the weight of the sample and the model's attention to sample detection. Thus, the model pays more attention to learning those samples that are difficult to detect.

On the basis of Focal Loss, Varifocal Loss proposes an asymmetric weighting operation, and its calculation formula is as follows:

$$\text{VFL} = \begin{cases} -q(q\log(\hat{y}) + (1-q)\log(1-\hat{y})) & q > 0 \\ -\alpha p_t^\gamma \log(1-\hat{y}) & q = 0 \end{cases} \tag{6}$$

When calculating the loss of positive samples, q > 0, its value is the Intersection over Union (IoU) of the prediction box and the labeling box; q = 0 means calculating the loss of negative samples.

It can be seen from Equation (6) that when the label is a positive sample, it is equivalent to using a BCE loss function with an adaptive IoU weighting value added. The IoU weighting is used to highlight the main samples (samples with high IoU value) among the positive samples. When the label is a negative sample, the function is the standard Focal Loss. This design not only balances the positive and negative samples but also highlights the main samples, ensuring that the training focuses on high-quality positive samples. We used Varifocal Loss as the confidence loss function of the model to alleviate the problem of sample imbalance.

*3.5. Slicing-Aided Hyper Inference (SAHI)*

SAHI is an open-source framework for detecting small objects in large-size images. Proposed by Akyon et al. [40], it can be mounted on any object detector. Figure 7 displays the working process of SAHI.

The large-size image to be detected is divided into several parts $P_1, P_2, ... , P_l$. The aspect ratio is maintained, and the size of each slice is adjusted. Then, each slice is put into the model for detection. The model detects the objects that exist in each slice and combines the detection results into a complete image. The method of merging the detection results of the sliced pictures is to map the position coordinates of the targets in all the sliced pictures to the original picture, so that the detection result of a complete large picture can be obtained. Finally, the Non-Maximum Suppression (NMS) algorithm is used to the remove duplicate prediction boxes.

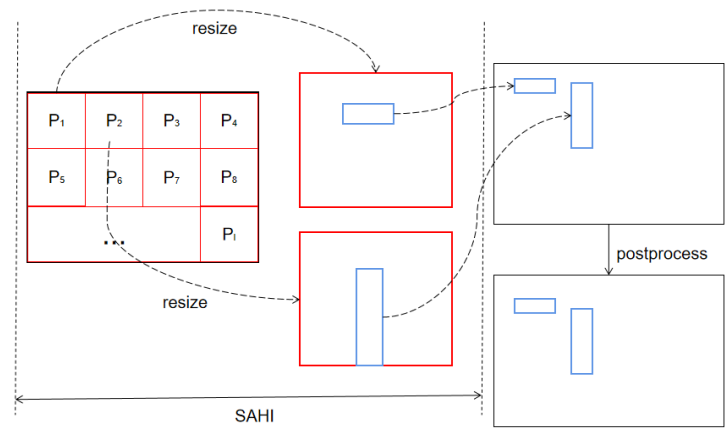

**Figure 7.** Execution process of the SAHI framework.

Object detection implementations in the SAHI framework can use any object detection algorithms. To subtly realize the object detection of high-resolution images, the framework uses image slicing and largely avoids the problem of missed detection and false detection of small targets in large-sized images. In practical applications, to avoid cutting off objects in the image, we performed overlapping cuts on the original image, and the duplicate detection frames obtained by mapping the target coordinates back to the original image could be filtered out in subsequent NMS processing. Satellite remote sensing images are usually high-resolution images, so SAHI is suitable for remote sensing image detection.

## 4. Experiments

*4.1. Experimental environment*

All experiments in this article were run in Python3.8, PyTorch1.10.0, and CUDA10.2. The CPU processor of the machine used was AMD Ryzen 5 3600X 6-Core, and the graphics card was NVIDIA GeForce RTX 2080 Ti. The speed and accuracy tests in the experiments were all conducted under the settings of FP16-precision and batch size = 1.

*4.2. Datasets*

The proposed RS-YOLOX model is evaluated on three aerial remote sensing image datasets: DOTA-v1.5 [42-44], TGRS-HRRSD [45, 46], and RSOD [47, 48].

**DOTA-v1.5** includes 2806 large size images; 16 categories; and 403,318 object instances. The DOTA images were collected from Google Earth, GF-2, and JL-1 satellite provided by the China Centre for Resources Satellite Data and Application. It is widely used in the research of remote sensing image object detection. The image size of this dataset is large (800×800 to 20,000×20,000 pixels), and the images could not be directly used for model training. So we cropped each image to a size of 640×640. Then we used data augmentation techniques to increase the number of small categories. The number of images after cropping and augmentation was 42,831. We split the images into a training set, a validation set, and a test set in a ratio of 6:2:2. The training set had 25,699 images, the validation set had 8566 images, and the test set had 8566 images.

**TGRS-HRRSD** includes 21,761 images; 13 categories; and 55,740 object instances. This dataset was acquired from Google Earth and Baidu Map. The image size of this dataset is from 150×150 to 1200×1200. The advantage of this dataset is that the number of categories is balanced and each category has about 4000 instances. The officially announced training set had 5401 images, validation set had 5417 images, and test set had 10,943 images.

**RSOD** includes 976 images, 4 categories, and 6950 object instances. The image size of this dataset is about 1000×1000. The number of different categories in this dataset is not balanced. To achieve better experimental results, we used data augmentation techniques to increase the number of images in small categories. The augmented dataset has a total of 1916 images, with approximately 400 images per category. We split the images into a training set, a validation set, and a test set in a ratio of 6:2:2. The training set had 1150 images, the validation set had 383 images, and the test set had 383 images.

*4.3. Evaluation metrics*

The performance of model in this article is evaluated comprehensively through AP, mAP, Latency, and FPS.

AP represents the average precision of a single category under different recall conditions, which is also equal to the area under the Precision–Recall curve. The mAP metric represents the mean of AP values of all categories and can be used to measure the accuracy of model detection. The formulas for calculating AP and mAP are as follows:

$$\mathbf{AP} = \int_0^1 \mathbf{P(r)dr} \tag{7}$$

$$\mathbf{mAP} = \frac{\sum_{k=1}^{n} \mathbf{AP^k}}{n} \tag{8}$$

$AP_{50}$ and $mAP_{50}$ represent the AP value and the mAP value, respectively, when the threshold of IoU is 0.50; $mAP_{50-95}$ is the average of 10 mAP values with IoU thresholds of 0.50, 0.55, 0.60, ..., 0.90, 0.95.

Latency represents the average inference time to detect an image; FPS means the number of image frames detected per second. Latency and FPS can be used to measure the detection speed of the model.

In addition, we used P to represent the total number of parameters of the model, which is used to measure the size of the model.

*4.4. Experiment results of proposed methods*

4.4.1. Effects of attention mechanisms

We added several advanced attention modules to the original YOLOX model for comparative experiments. SE [26] is a classic channel attention mechanism, CBAM [27] combines channel attention and spatial attention, NAM [49] is a lightweight mixed attention module, which redesigns the implementation of channel attention and spatial attention in CBAM, ULSAM [50] is a lightweight subspace attention module, and ECA [37] is an improved version of SE. We added these attention mechanisms to the back of the backbone network so that the features processed by the attention mechanisms could be sent to PAFPN. We trained the model on the processed DOTA-v1.5 dataset and obtained the results shown in Table 1.

**Table 1.** Test results of attention mechanism blocks on DOTA.

| Methods | mAP$_{50\text{-}95}$ (%) | P(M) | Latency (ms) | FPS |
|---|---|---|---|---|
| YOLOX | 73.57 | 8.94 | 13.66 | 73.2 |
| YOLOX+SE | 74.11 | 9.06 | 14.12 | 70.8 |
| YOLOX+CBAM | 73.98 | 9.15 | 15.89 | 62.9 |
| YOLOX+NAM | 74.21 | 8.95 | 15.45 | 64.7 |
| YOLOX+ULSAM | 74.06 | 8.95 | 17.94 | 55.8 |
| YOLOX+ECA | **75.58** | 8.94 | 13.26 | 75.4 |

It can be seen from Table 1 that the effect of ECA is the best among these attention mechanisms, the number of parameters added is the least, and the accuracy of the model can be improved without increasing the number of parameters. Thus, we finally chose the ECA attention mechanism to add to the YOLOX model.

The results of our testing of the ECA module using three datasets are displayed in Table 2. After we used the ECA module, the detection accuracy of the model when used on these three datasets was improved and the speed of image detection was also slightly accelerated.

**Table 2.** Effects of ECA on three datasets.

| Methods | DOTA-v1.5 | | TGRS-HRRSD | | RSOD | |
|---|---|---|---|---|---|---|
| | mAP$_{50\text{-}95}$ (%) | Latency (ms) | mAP$_{50\text{-}95}$ (%) | Latency (ms) | mAP$_{50\text{-}95}$ (%) | Latency (ms) |
| YOLOX | 73.57 | 13.66 | 66.07 | 11.34 | 77.18 | 14.86 |
| YOLOX+ECA | 75.58 | 13.26 | 69.23 | 11.21 | 78.23 | 14.58 |

4.4.2. Effects of the ASFF algorithm

After adding ASFF [38] to the neck network in the YOLOX model, we obtained the results shown in Table 3.

We tested ASFF on three datasets, and the accuracy of the model was greatly improved. However, the parameters and calculations increased and the time of model forward inference increased. Although it increases the number of parameters and computation to a certain extent, it is also a feasible method to improve the accuracy without the need to achieve real-time detection.

**Table 3.** ASFF test results.

| Methods | P(M) | DOTA-v1.5 | | TGRS-HRRSD | | RSOD | |
|---|---|---|---|---|---|---|---|
| | | mAP$_{50\text{-}95}$ (%) | Latency (ms) | mAP$_{50\text{-}95}$ (%) | Latency (ms) | mAP$_{50\text{-}95}$ (%) | Latency (ms) |
| YOLOX | 8.94 | 73.57 | 13.66 | 66.07 | 11.34 | 77.18 | 14.86 |
| YOLOX+ASFF | 14.38 | 75.62 | 14.38 | 66.92 | 13.97 | 78.01 | 18.60 |

#### 4.4.3. Loss function test

The confidence loss function used by the YOLOX model is BCE, and although it is a classic and effective classification loss function, it did not balance the positive and negative samples in training. Therefore, we replaced it with Focal Loss (FL) [41] and Varifocal Loss (VFL) [39] for experiments and obtained the results shown in Table 4. As per Table 4, among the three loss functions, VFL can obtain the highest detection accuracy and reduces the inference time compared to FL. Therefore, we finally chose to use VFL as our confidence loss function.

**Table 4.** Test results of different confidence loss functions.

| Methods | DOTA-v1.5 | | TGRS-HRRSD | | RSOD | |
|---|---|---|---|---|---|---|
| | mAP$_{50\text{-}95}$ (%) | Latency (ms) | mAP$_{50\text{-}95}$ (%) | Latency (ms) | mAP$_{50\text{-}95}$ (%) | Latency (ms) |
| YOLOX–BCE | 73.57 | **13.66** | 66.07 | **11.34** | 77.18 | **14.86** |
| YOLOX–FL | 75.08 | 15.91 | 66.92 | 13.18 | 77.46 | 16.53 |
| YOLOX–VFL | **75.52** | 14.04 | **67.13** | 11.87 | **78.03** | 15.27 |

#### 4.4.4. Effects of SAHI

Because our experiments were trained using cut remote sensing images, when actually applied to large-sized remote sensing images, we could still detect targets only after the images were cut. When we used our trained model, Figure 8 displays the detection results of the remote sensing images after the images were cut.

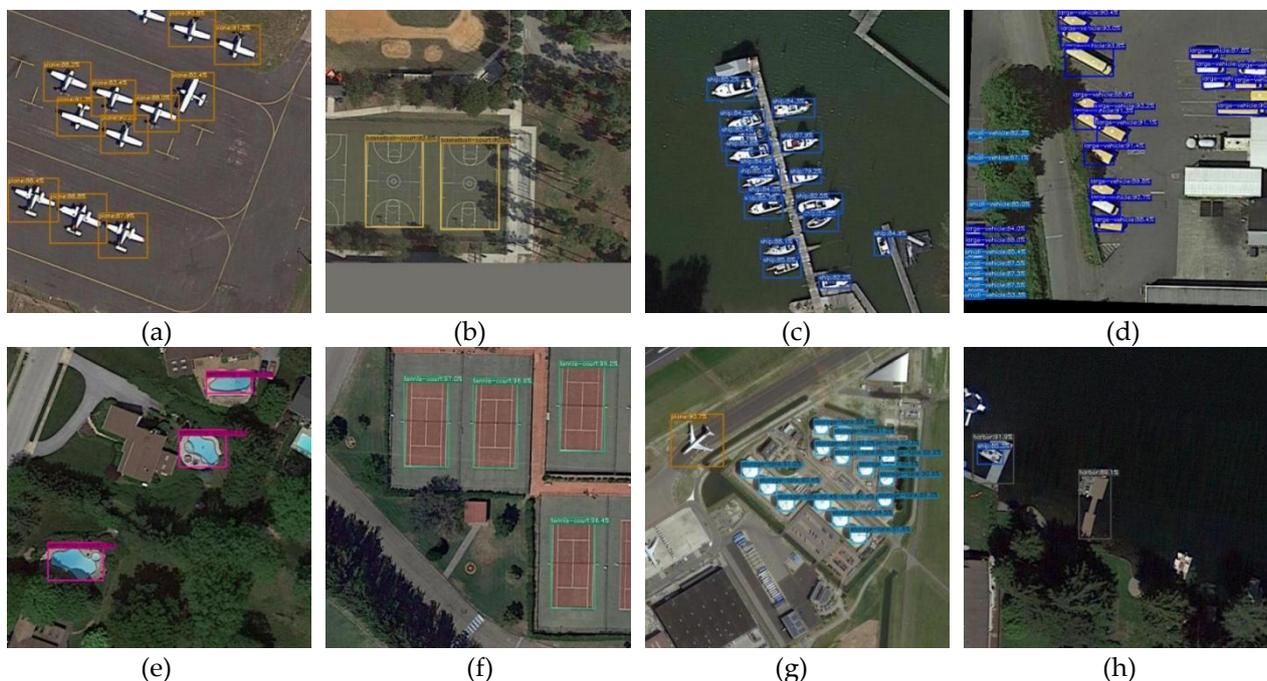

**Figure 8.** Detection results of small images on DOTA-v1.5 after the images were cut. (a) Plane, (b) basketball court, (c) ship, (d) large vehicle, (e) swimming pool, (f) tennis court, (g) storage tank, and (h) harbor.

Our model performed well on these small images in Figure 8, detecting all the objects in these images. However, if we directly use the model to detect a high-resolution image, the detection effect will be poor, as shown in Figure 9. This is the detection result of a high-resolution image (a size of 3826×3473). A large number of objects are not de-tected, and there are many false detections. In response to this problem, we introduced the SAHI [40] open-source framework and mounted our trained YOLOX model onto the SAHI framework. The detection result obtained is shown in Figure 10. It can be clearly seen that the use of SAHI greatly improves the detection performance of the model for large-size images.

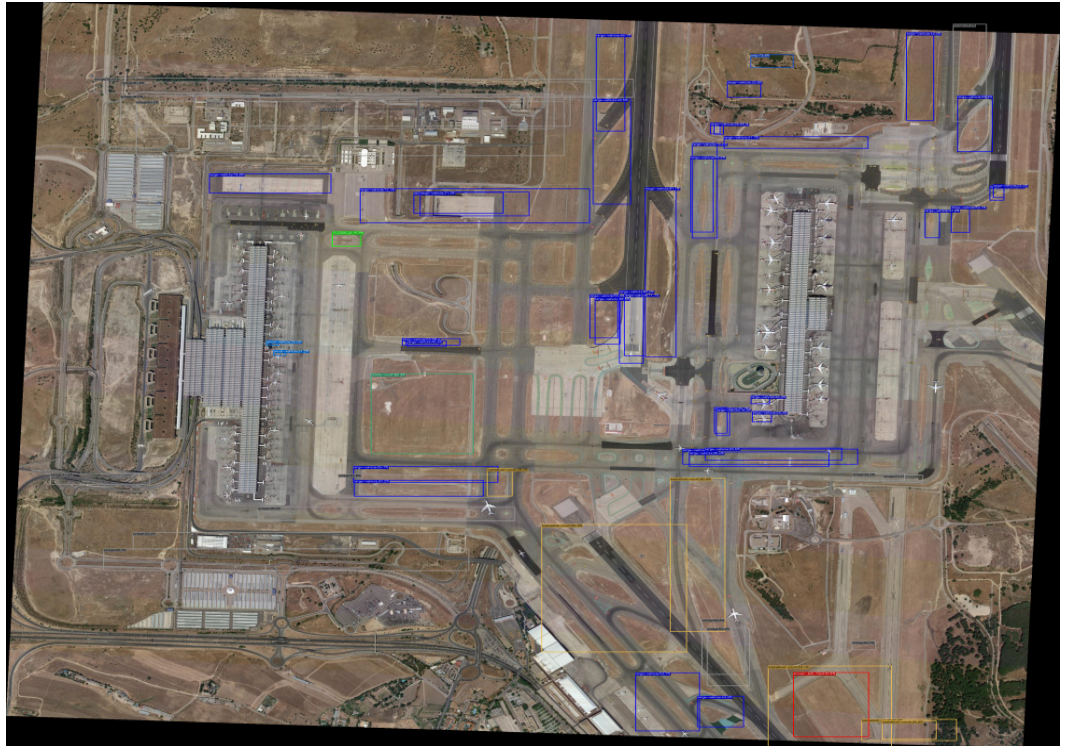

**Figure 9.** The detection result of putting a large-size image directly into the YOLOX mode.

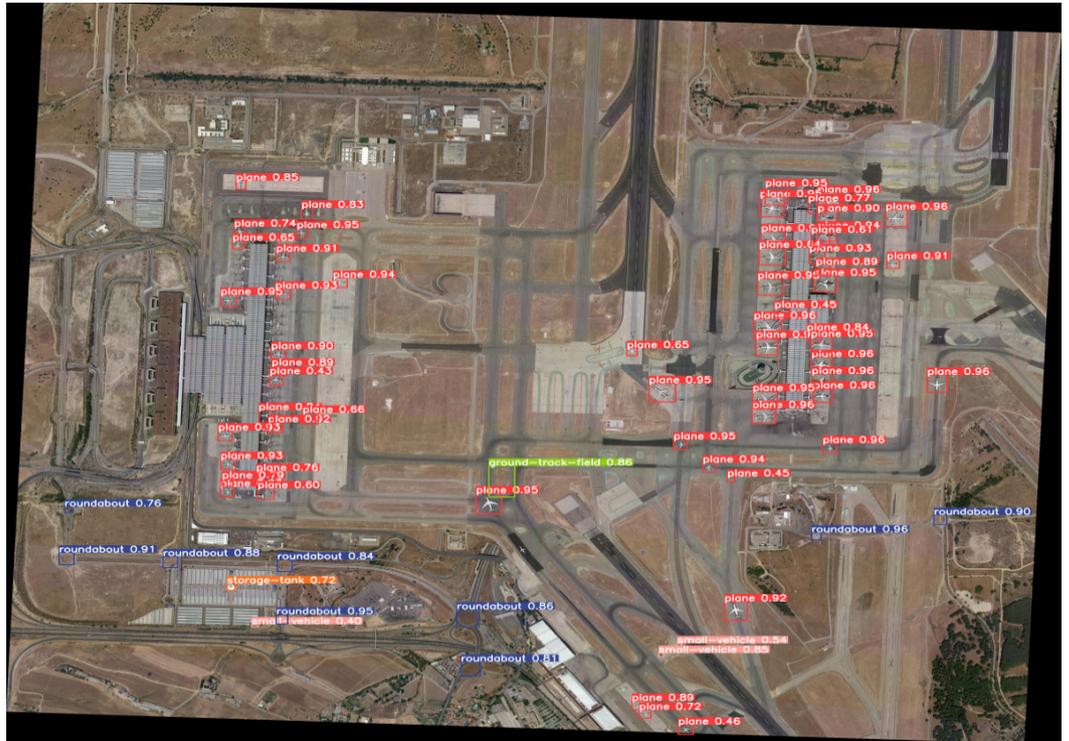

**Figure 10.** The detection results of the YOLOX model combined with the SAHI framework for the same high-resolution image.

*4.5. Evaluation model*

We added the above improved methods to the original YOLOX model in turn and tested them with the cut and amplified DOTA-v1.5 dataset. The results obtained are displayed in Table 5.

**Table 5.** The test results of adding each module to YOLOX in turn on DOTA-v1.5.

| Methods | mAP$_{50\text{-}95}$ (%) | P(M) | Latency (ms) | FPS |
|---|---|---|---|---|
| YOLOX | 73.57 | 8.94 | 13.66 | 73.2 |
| +ECA | 75.58(+2.01) | 8.94 | 14.46 | 69.2 |
| +ASFF | 77.16(+1.58) | 14.38 | 17.96 | 55.7 |
| +VFL | **78.49(+1.33)** | 14.38 | 17.67 | 56.6 |

After we improved the YOLOX model, the test results on the DOTA dataset were 4.92% higher than the mAP of the original YOLOX model, but there was a slight increase in the number of parameters and calculations. Finally, we fused the improved YOLOX and SAHI frameworks so that our model could directly detect large-size remote sensing images. By improving YOLOX using all the above methods, we produced our new model: RS-YOLOX.

It can be seen from the experimental results that, compared with YOLOX, RS-YOLOX greatly improved the detection performance for aerial remote sensing images with extremely high resolution. Figure 11 shows a comparison of object detection results of YOLOX and RS-YOLOX on the DOTA-v1.5 dataset.

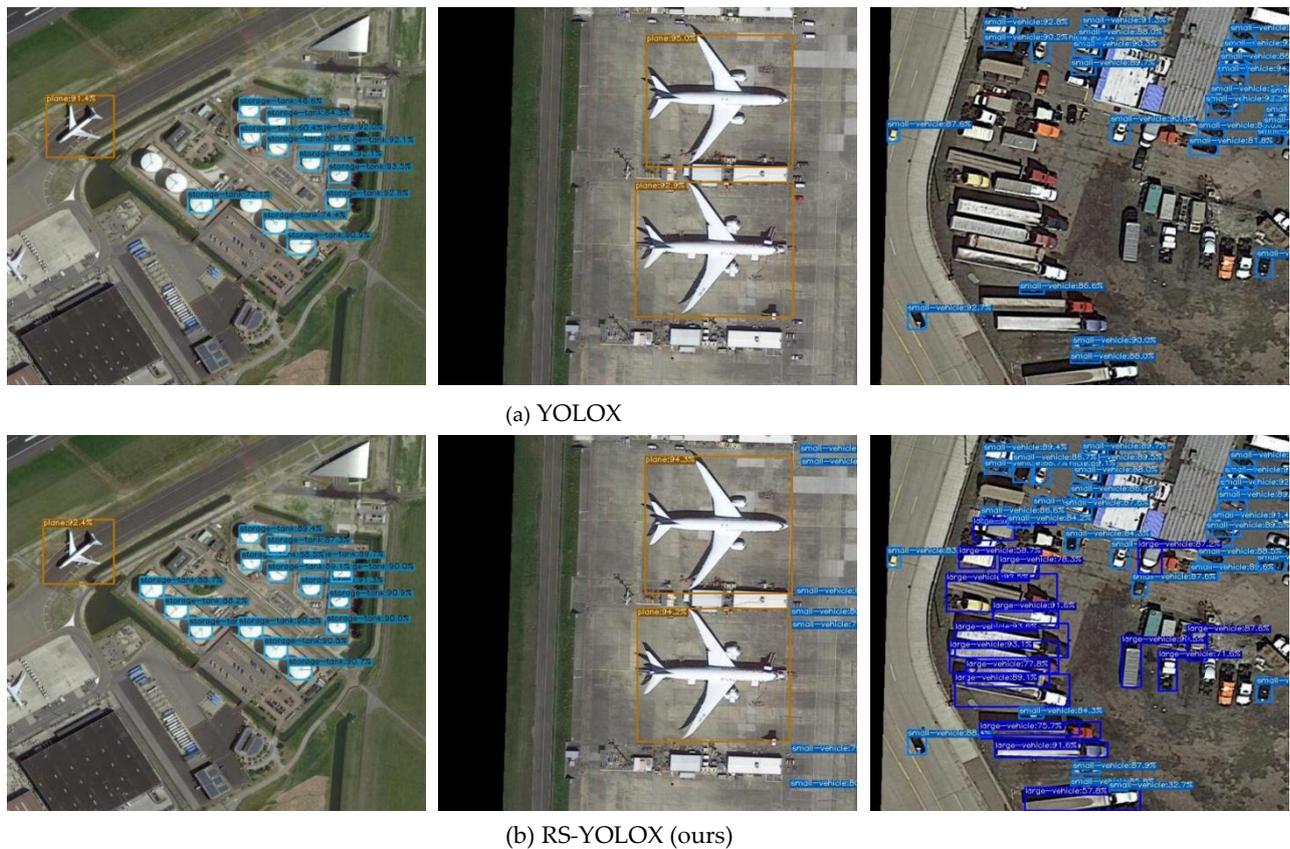

(a) YOLOX

(b) RS-YOLOX (ours)

**Figure 11.** Comparison of detection results of YOLOX and RS-YOLOX on the DOTA-v1.5 dataset.

Our model also achieved good performance on two other datasets. Figure 12 displays its detection results on TGRS-HRRSD and RSOD datasets. The official website of TGRS-HRRSD provides the detection results of several advanced models. Table 6 presents a comparison between our model and these models.

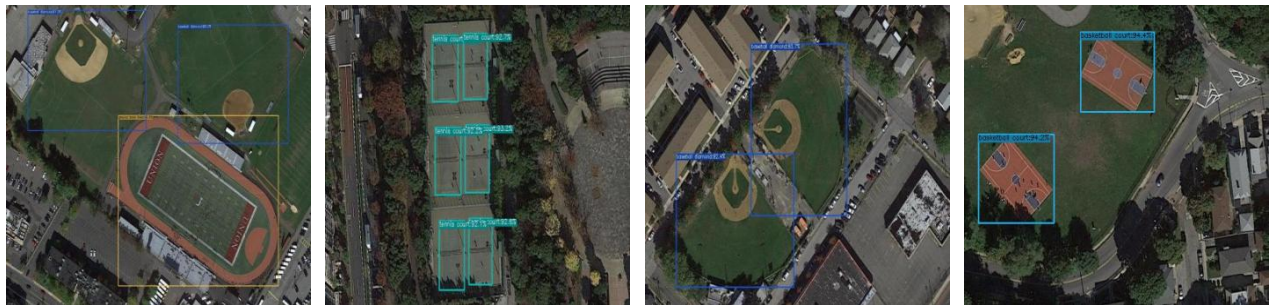
(a) Detection results of our model on TGRS-HRRSD.

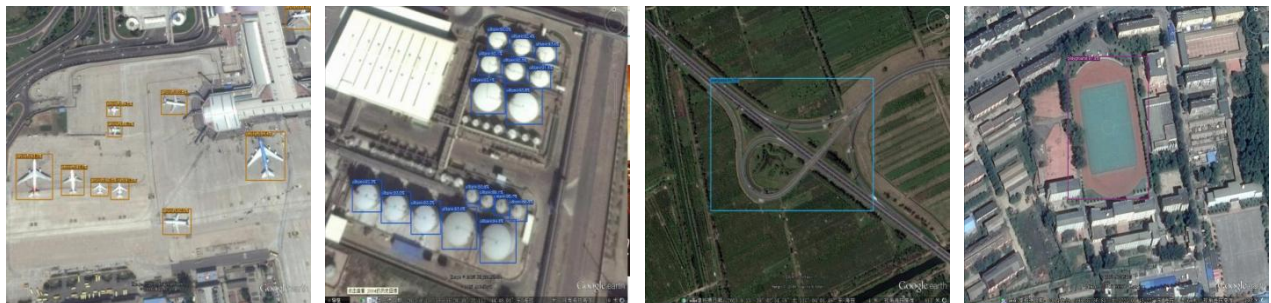
(b) Detection results of our model on RSOD.

**Figure 12.** Detection results of our model on TGRS-HRRSD and RSOD. (a) From left to right, the photographs show ground track fields, tennis courts, baseball diamonds, and basketball courts. (b) From left to right, the photographs show aircraft, oil tanks, an overpass, and a playground.

**Table 6.** Comparison of our model with the models on the HRRSD official website.

| Category | YOLOv2 /% | Fast R-CNN | Fast_R-CNN+ GACL-Net /% | Faster R-CNN | Faster_R-CNN+ GACL-Net /% | RS-YOLOX (ours) /% |
|---|---|---|---|---|---|---|
| Airplane | 84.6 | 83.3 | 85.1 | 90.8 | 90.8 | **90.9** |
| Baseball Diamond | 62.2 | 83.6 | 82.6 | 86.9 | 87.2 | **88.6** |
| Basketball Court | 41.3 | 36.7 | 42.1 | 47.9 | 49.7 | **78.2** |
| Bridge | 79.0 | 75.1 | 76.7 | 85.5 | 85.6 | **90.4** |
| Crossroad | 43.4 | 67.1 | 68.7 | 88.6 | 88.2 | **90.1** |
| Ground Track Field | 94.4 | 90.0 | 89.6 | 90.6 | 90.7 | **90.8** |
| Harbor | 74.4 | 76.0 | 78.4 | 89.4 | 89.7 | **89.9** |
| Parking Lot | 45.8 | 37.5 | 39.5 | 63.3 | 65.3 | **75.0** |
| Ship | 78.5 | 75.0 | 74.3 | 88.5 | 88.5 | **90.1** |
| Storage Tank | 72.4 | 79.8 | 80.4 | 88.7 | 89.2 | **90.3** |
| T Junction | 46.8 | 39.2 | 38.8 | 75.1 | 75.0 | **86.5** |
| Tennis Court | 67.6 | 75.0 | 77.0 | 80.7 | 80.0 | **90.4** |
| Vehicle | 65.1 | 46.1 | 50.7 | 84.0 | 86.9 | **90.8** |
| mAP$_{50}$ | 65.8 | 66.5 | 68.0 | 81.5 | 82.1 | **88.0** |

On the TGRS-HRRSD dataset, our model has the highest mAP value and our model has a higher detection accuracy for each category than the models provided by the official website.

Finally, we tested the current state-of-the-art remote sensing image detectors with these three datasets, and the results obtained are displayed in Table 7.

**Table 7** Performance comparison of advanced object detection models on aerial remote sensing images.

| Dataset | Evaluation Metrics | Improved Faster R-CNN [21] | YOLT [23] | YOLO v3 [17] | YOLO v4 [18] | YOLOv5 [19, 24] | YOLOX [20] | RS-YOLOX (ours) |
|---|---|---|---|---|---|---|---|---|
| DOTA-v1.5 | $mAP_{50}$ (%) | 72.51 | 72.06 | 77.05 | 86.23 | 88.56 | 90.12 | **92.83** |
| | $mAP_{50-95}$ (%) | 50.26 | 51.27 | 58.32 | 69.01 | 71.38 | 73.57 | **78.49** |
| | Latency (ms) | 263.29 | 25.68 | 19.11 | 16.13 | **11.87** | 13.66 | 17.86 |
| | P(M) | 60.42 | 50.54 | 61.60 | 64.02 | **7.05** | 8.94 | 14.38 |
| TGRS-HRRSD | $mAP_{50}$ (%) | 74.29 | 73.19 | 74.38 | 85.63 | 85.87 | 86.01 | **88.04** |
| | $mAP_{50-95}$ (%) | 53.23 | 54.17 | 55.26 | 64.58 | 63.33 | 65.12 | **68.26** |
| | Latency (ms) | 260.78 | 23.96 | 18.12 | 15.02 | **9.24** | 11.34 | 15.23 |
| | P(M) | 60.42 | 50.54 | 61.60 | 64.2 | **7.50** | 8.94 | 14.38 |
| RSOD | $mAP_{50}$ (%) | 80.79 | 79.28 | 81.36 | 90.23 | 90.97 | 92.01 | **93.07** |
| | $mAP_{50-95}$ (%) | 62.12 | 61.33 | 64.57 | 76.12 | 75.34 | 76.27 | **78.56** |
| | Latency (ms) | 265.35 | 28.72 | 21.79 | 17.56 | **11.88** | 14.86 | 18.29 |
| | P(M) | 60.42 | 50.54 | 61.60 | 64.2 | **7.50** | 8.94 | 14.38 |

Our model achieves good results on three aerial image datasets, and its detection accuracy surpasses all of the above models.

## 5. Discussion

This section discusses the links between previous research and ours, the strengths and weaknesses of our model, and our future directions for improvement. The experiments in Section 4.4 show that all the methods we used have good effects on improving the accuracy of remote sensing image detection models. The experiments in Section 4.5 show that our model has the highest detection accuracy compared to other state-of-the-art models. Before us, many researchers have also tried to use a combination of various advanced methods to achieve automatic object detection in remote sensing.

### 5.1. About the backbone network

To improve the quality of features extracted by the backbone network, some researchers have tried to add an attention mechanism to the backbone network. Li et al. [51] combined the resnet-50 and the position attention module (PAM) to achieve high accuracy in the change detection of remote sensing images. Zhang et al. [52] introduced a new spatial attention mechanism into the Siamese network, which also improved the accuracy of the model for detecting changes in multispectral images. Our research drew on these previous studies, and added the ECA attention mechanism after the backbone network of YOLOX, which improved the detection accuracy of the model.

To find the most suitable attention mechanism, we compared the results of five attention mechanisms (as shown in Table 1), which include three types of attention: spatial, channel, and mixed. From the results of detection accuracy and speed, ECA has the best results. We found that channel domain attention is the most friendly to our model and dataset, it improves accuracy without adding excessive inference time.

Previous researchers used some lightweight networks to replace the backbone network in order to realize the lightweight of the model. For example, Li et al. [53] replaced the backbone network of the YOLO model with Mobilenet, which greatly reduced the amount of parameters and calculations, and realized real-time detection of remote sensing images. This is a general way to improve the detection speed of the model. However, the Mobilenet constructed by depthwise separable convolution has low detection accuracy, so our research did not use this method.

*5.2. About the neck network*

The neck network is usually used to achieve feature fusion. Researchers have tried to improve the feature fusion of the neck network to improve the performance of the model. Yan et al. [21] used the feature pyramid network (FPN) in Faster RCNN to achieve multi-scale feature fusion, and used channel attention in the FPN structure to achieve high accuracy in detecting tailing ponds in remote sensing images. Qu et al. [54] introduced the ASFF structure in the FPN of YOLOv3, which strengthened the feature fusion of FPN, and achieved good detection results.

Besides FPN, PAFPN used by YOLOX is also a commonly used neck network, so we tried to connect the ASFF structure after the neck network. The results are shown in Table 3, and the accuracy of the model has improved significantly. However, a disadvantage of ASFF is also very obvious. It increases the amount of parameters and calculation of the model.

*5.3. About the loss function*

YOLOX uses the BCE function as the classification and confidence loss function, and the IoU loss as the positioning loss function. Some studies [39, 41] have shown that using a focal loss type of loss function is better than using the BCE function. Because focal loss can alleviate the imbalance in the number of positive and negative samples. Based on these theoretical foundations, we replaced the BCE function in YOLOX to the latest focal loss function, Varifocal Loss, and obtained the accuracy improvement (as shown in Table 4).

We have also studied the replacement of the positioning loss function. In addition to the IoU function, the positioning loss functions include GIoU [55], DIoU [56], CIoU [56] and SIoU [57]. They all add other punishment items on the basis of IoU to obtain a better convergence effect. We tried to replace the IoU function in YOLOX into these functions, but the improvement effect of the detection accuracy was not obvious, and the reasoning time was increased. So we did not use these more advanced loss functions in our final model.

*5.4. About our model and future improvement*

Judging from the experimental results, our model undoubtedly has high detection accuracy, and with the use of the SAHI framework, our model can directly detect large-sized remote sensing images. However, compared to the original model, our improved YOLOX model has more parameters and calculations, so the detection speed has decreased. The results of the ablation experiments in Table 5 can be known that the increased parameters and calculations mainly comes from the ASFF structure we used.

For the problem of detection speed drop, our future research is mainly aimed at improving the detection speed, we want to improve our detection speed while maintaining the high accuracy of the model. We consider redesigning the structure of ASFF, reducing its number of parameters, or trying other variants of the FPN structure. We will also consider deploying our model using TensorRT to reduce the inference time of the model.

*5.5. Other interesting discussions*

Our research is to use deep learning technology to achieve automatic detection of objects in remote sensing images. In the introduction part, we mentioned that the method of manually extracting features is not enough to fully extract the target features, and using deep learning methods to automatically extract features can greatly improve the efficiency.

In fact, in the field of remote sensing applications, although automated methods are highly efficient, they may not be the optimal method in certain specific environments. In some cases, manual methods of processing remote sensing data can also achieve good results. For example, Weißmann et al. [58] collected data by controlling the drone, and then used related softwares to reconstruct the point clouds and textured 3D models of the survey objects. This study did not use a fully automated process, and many critical steps were performed manually, but according to these methods, photorealistic 3D models can be reconstructed. This shows that in the field of remote sensing, the use of artificial methods can also achieve good enough results.

Automated methods can process large amounts of data in a short period of time, they have the advantages of convenience and high efficiency. However, they cannot replace the flexible thinking of humans. In some situations that require complex operations, manual methods are obviously more reliable than automated methods. We pursue technological innovation to facilitate people's lives, but we do not want the automation of machines to replace some activities that require human creativity. Because the accumulation of human experience and knowledge is the most valuable treasure, which cannot be replaced by machines.

**6. Conclusion**

In this article, for the detection task of aerial remote sensing images, we made a series of improvements to the YOLOX model, using three aerial remote sensing datasets for training and testing. We finally proposed an improved, high-performance remote sensing image detection model: RS-YOLOX.

To enhance the features in the backbone network and the neck network and improve the performance of the model in detecting small targets, we used ECA and ASFF. To adjust the number of positive and negative samples, we used the VFL function. Finally, to improve the detection performance of the model for large-size images, we adapted the model to the SAHI framework. Although our model has high detection accuracy for remote sensing images, compared with the original YOLOX model, it increases the parameters and calculation as well as the detection time. Therefore, our next plan is to study the model in terms of making it lightweight and improving its speed.

**Author Contributions:** L.Y. proposed the network architecture design and the framework of RS-YOLOX. H.L. and J.C. collected and preprocessed the datasets. L.Y. performed the experiments. L.Y., H.L., and J.C. analyzed and discussed the experimental data. L.Y. and G.Y. wrote the article. G.Y., H.Z. and H.W. revised the article and provided valuable advice for the experiments. All authors have read and agreed to the published version of manuscript.

**Funding:** This research was funded by the Key R&D projects in Yunnan Province (Grant No. 202202AD080004), the Application and Foundation Project of Yunnan Province (Grant No.202001BB050032), the Yunnan Provincial Department of Science and Technology-Yunnan University Joint Special Project for Double-Class Construction (Grant No. 202201BF070001-005) and the Open Project of CAS Key Laboratory of Solar Activity, National Astronomical Observatories (Grant No. KLSA202115).

**Data Availability Statement:** The three datasets that support this study are openly available online.
DOTA-v1.5: https://captain-whu.github.io/DOTA/dataset.html,
TGRS-HRRSD: https://github.com/CrazyStoneonRoad/TGRS-HRRSD-Dataset,
RSOD: https://github.com/RSIA-LIESMARS-WHU/RSOD-Dataset-.

**Acknowledgments:** We would like to thank the anonymous reviewers and the editor-in-chief for their comments to improve the article. Thanks also to the data sharer. We thank all the people in-volved in the study.